\documentclass{article}
\usepackage{amsmath,graphicx,mlspconf}
\usepackage{booktabs}
\usepackage{amssymb}
\usepackage{mathtools}
\usepackage{hyperref}

\copyrightnotice{U.S.\ Government work not protected by U.S.\ copyright}

\copyrightnotice{979-8-3503-2411-2/23/\$31.00 {\copyright}2023 Crown}

\copyrightnotice{979-8-3503-2411-2/23/\$31.00 {\copyright}2023 European Union}

\copyrightnotice{979-8-3503-2411-2/23/\$31.00 {\copyright}2023 IEEE}

\toappear{2023 IEEE International Workshop on Machine Learning for Signal Processing, Sept.\ 17--20, 2023, Rome, Italy}

\def\x{{\boldsymbol{x}}}
\def\h{{\boldsymbol{h}}}
\def\z{{\boldsymbol{z}}}
\def\g{{\boldsymbol{g}}}

\DeclarePairedDelimiter\parentheses{\lparen}{\rparen}

\newcommand{\cossit}[4]{\operatorname{s}_{{\tau \parentheses*{#1,#2}}}^{\parentheses*{#3,#4}}}

\newcommand{\fexp}[1]{\exp\parentheses*{#1}}

\newcommand{\encoder}[1]{\operatorname{H}_{\boldsymbol{\theta}}\parentheses*{#1}}

\title{Multi-view self-supervised learning for multivariate variable-channel time series} 
\name{Thea Brüsch, Mikkel N. Schmidt, Tommy S. Alstrøm}
\address{Department of Applied Mathematics and Computer Science, Technical University of Denmark}

\begin{document}

\maketitle

\begin{abstract}
Labeling of multivariate biomedical time series data is a laborious and expensive process. Self-supervised contrastive learning alleviates the need for large, labeled datasets through pretraining on unlabeled data. However, for multivariate time series data, the set of input channels often varies between applications, and most existing work does not allow for transfer between datasets with different sets of input channels. We propose learning one encoder to operate on all input channels individually. We then use a message passing neural network to extract a single representation across channels. We demonstrate the potential of this method by pretraining our model on a dataset with six EEG channels and then fine-tuning it on a dataset with two different EEG channels. We compare models with and without the message passing neural network across different contrastive loss functions. We show that our method, combined with the TS2Vec loss, outperforms all other methods in most settings. 


\end{abstract}
\section{Introduction}
\label{sec:intro}
In recent years, self-supervised learning has shown promising results in the fields of computer vision and natural language processing~\cite{simclr, bert}. Self-supervised learning relies on inherent patterns within the data to enable pretraining on large, unlabeled datasets, thus facilitating the transfer of learned structures to smaller labeled datasets, usually called the downstream tasks. Obtaining ground truth scoring for biomedical signals such as electroencephalography (EEG) often requires the expertise of multiple professionals, rendering label acquisition a challenging and expensive endeavor~\cite{younes2017a}. Consequently, self-supervised learning methods are particularly interesting for biomedical time series data.

Many self-supervised learning methods use \emph{contrastive learning} to pretrain the networks. Contrastive learning relies on having both positive and negative pairs, where the positive pairs are encouraged to be close and the negative pairs distant in representation space~\cite{1640964}. Non-contrastive self-supervised learning tasks include the reconstruction of masked input pixels and loss functions that only require positive views. In this work, we focus on \emph{contrastive self-supervised learning}. 


Previous work on contrastive pretraining for time series data uses various different strategies to create positive pairs. Broadly speaking, we divide the strategies into three categories. The first category uses augmentations such as masking, scaling, or random additive noise. The second category uses contrastive predictive coding (CPC), where an autoregressive model is used to predict future samples. A closely related strategy uses a combination of masking and CPC to reconstruct masked out segments within the current sequence. The third category relies on data that inherently contains multiple views, such as multiple channels or different modalities. We refer to the third strategy as a multi-view strategy. 

Previous significant work on contrastive pretraining for time series data includes Eldele et al.~\cite{ijcai2021-324}, who use augmentations such as permutations and scaling. Furthermore, they use a temporal contrasting strategy similar to CPC to predict future augmented samples. Zhang et al.~\cite{zhang2022selfsupervised} use similar augmentations but create a separate encoder in the frequency domain and encourage time and frequency representations to be close. Yue et al.~\cite{TS2Vec} use random cropping and masking to augment the input signal as well as a new hierarchical time series loss to train their model, which they call TS2Vec. 
BErt-like Neurophysiological Data Representation (BENDR) by Kostas et al.~\cite{BENDR} comprises a convolutional encoder that tokenizes raw input EEG, and a transformer that contextualizes the tokens. The network is then trained using a combination of CPC and masking. 
Kiyasseh et al.~\cite{kiyasseh2021clocs} and Deldari et al.~\cite{cocoa} both leverage the multi-view strategy for creating positive pairs.
Kiyasseh et al.~\cite{kiyasseh2021clocs} investigate contrastive pretraining for electrocardiography (ECG). They use both neighboring samples in time and different channels as positive pairs. Finally, Deldari et al.~\cite{cocoa} use different sensor modalities as positive pairs and present a new loss, COCOA, tailored for contrastive learning in settings with more than one view. We focus our work on the multi-view strategy for multivariate time series data. 

A significant challenge for self-supervised learning applied on multivariate time series is that the number of channels may vary from application to application. The varying number of channels makes it difficult to transfer between tasks with different channels~\cite{9492125}, and few of the current methods have a principled way of handling this issue. The mentioned previous work either pretrain and fine-tune on the same dataset~\cite{ijcai2021-324, TS2Vec, cocoa}, or discard excess channels or zero-pad missing channels during fine-tuning and/or pretraining~\cite{zhang2022selfsupervised, BENDR}. The work most closely related to ours is SeqCLR by Mohsenvand et al.~\cite{seqclr}. SeqCLR is a single encoder that works separately on all channels individually. The encoder is pretrained using augmentations. During fine-tuning, the outputs of all input channels are concatenated and used as input to the classifier. 

We propose a channel agnostic network that generalizes between datasets with varying sets of input channels with no further preprocessing. We learn a single-channel encoder and add a message passing neural network (MPNN) after the encoder to extract the optimal combination of the individual channel representations. We use the different channels of the multivariate time series to create the positive pairs during pretraining. 
We demonstrate the use of the MPNN by pretraining on an EEG dataset with six channels and fine-tuning on an EEG dataset with two different channels, and compare different loss functions in the pretraining phase. Our results show that when combined with the TS2Vec loss, our method outperforms all other methods on most sample sizes.

\section{Methods}

\subsection{Channel agnostic setup}
\begin{figure}[t]
    \centering
    \includegraphics[width=\linewidth]{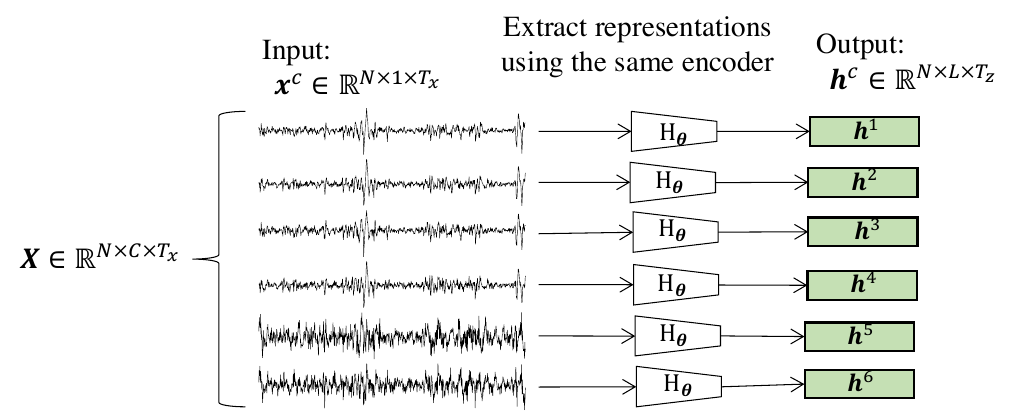}
    \caption{We apply the same encoder $\operatorname{H}_{\boldsymbol{\theta}}$ to each of the $C$ input channels $\x^{c}$ to obtain $C$ representations $\h^{c}$. Here, the setup is shown for $C=6$ channels.}
    \label{fig:encoder}
\end{figure}

We use a convolutional encoder to extract representations from the raw EEG signals. Our encoder follows the architecture in BENDR~\cite{BENDR} with the exception that we take only one channel as input. Given an input $\boldsymbol{X} \in \mathbb{R}^{N \times C \times T_{\text{in}}}$ with $N$ samples of raw EEG with $C$ channels, and each with a length of $T_{\text{in}}$, we take each channel, $\x^{c}$, and apply the same encoder $\operatorname{H}_{\boldsymbol{\theta}}$ to obtain the representation $\h^{c}$:
\begin{equation}
    \h^{c} = \encoder{\x^{c}}, \quad  \x^{c} \in \mathbb{R}^{N \times 1 \times T_{\text{in}}}, \quad \h^{c} \in \mathbb{R}^{N \times L \times T_{\text{\text{out}}}},
\end{equation}
where $L$ is the output dimension of the encoder and $T_{\text{out}}$ is the length after downsampling in the encoder. This setup is visualized in \autoref{fig:encoder} for $C=6$ channels.
Based on these $C$ representations, we use two different methods for creating the different positive views $\z^{v}$ for contrastive learning. 

In the first approach, we simply use each representation $\h^{c}$ for each view:
\begin{equation}
    \z^{v} = \h^{c}, \quad (c,v) \in \{(1,1),\dots, (C,C)\}.
\end{equation}
This approach results in $V$ positive views, $\z^{v}\in \mathbb{R}^{N \times L \times T_{\text{out}}}$. The $V$ positive views form $C(C-1)/2$ positive pairs per datapoint to use for contrastive learning. 

When fine-tuning on a downstream task with $C^{d}$ channels, this approach produces $C^{d}$ representations, $[\z^1,\dots,\z^{C^d}]$. To obtain a single representation across all channels, i.e., to use for classification, we add a linear layer of size $C^{d}\times 1$, which combines the $C^d$ representations into one representation. The linear layer is optimized during fine-tuning. 

\begin{figure}[t]
    \centering
    \includegraphics[width = \linewidth]{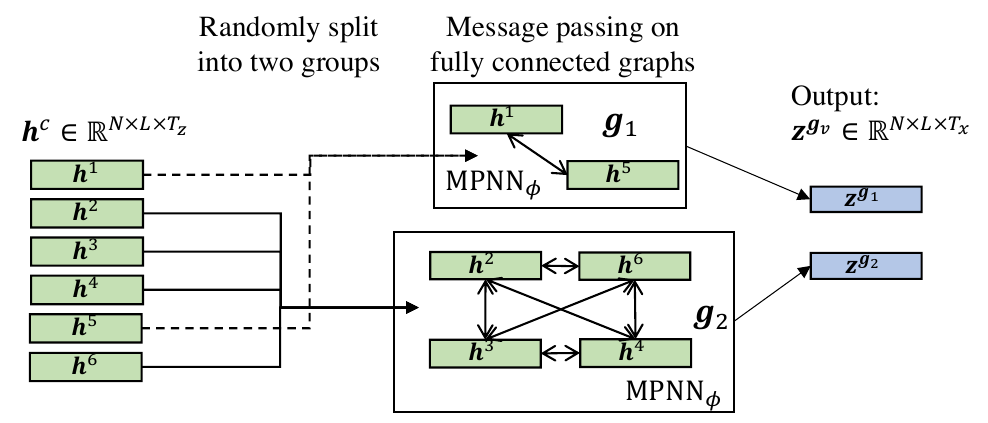}
    \caption{In our second multi-view setting, we split the $C$ representations into two random groups consisting of at least two channels each. Here, group 1 consists of $C_{1}=2$ randomly chosen representations and group 2 ccomprises the remaining $C_{2}=4$ representations. 
    }
    \label{fig:mpnn}
\end{figure}

In the second approach, for each new batch, we randomly divide the $C$ representations into two groups, $\g_1$ and $\g_2$, with $C_1$, and $C_2 = C-C_1$ representations respectively, where $C_1, C_2 \geq 2$. \autoref{fig:mpnn} shows an example partitioning. The partitioning (both exact split and partition size) of the representations is chosen randomly for each batch. 

In both groups, we form a fully connected graph and use identical message passing neural networks (MPNN) to extract the intra-group context for both groups individually:
\begin{equation}
\label{eq:mpnn_overview}
    \z^{v} = \z^{g_v} = \operatorname{MPNN}_{\phi}(\g_v), v \in [1,2].
\end{equation}
This approach results in $V=2$ positive views, $\z^{v}\in \mathbb{R}^{N \times L \times T_{\text{out}}}$. The two positive views form one positive pair for each data point to use for contrastive learning.

\subsection{Message passing neural network}
As stated in eq. \eqref{eq:mpnn_overview}, the MPNN is used to extract the intra-group context for each of the two groups. MPNNs were originally formalized in \cite{mpnn} and we follow their definition. An MPNN acts on graphs, and in our case, we form a fully connected directed graph within both of the groups, $\g_v$. This means that the input graph consists of $C_v$ vertices with vertex features $\h$. 

The MPNN consists of two phases, the message passing phase, and the readout phase. The message passing phase takes place in $K$ rounds, defined by the message passing networks $\operatorname{M}_{\phi_{k}}$, and an update operation $\operatorname{U}_{k}$. In each round, we compute the message and update the state for all $\h$ in $\g_v$:
\begin{equation}
\begin{aligned}
\boldsymbol{m}_{k+1}^{\h} & =\frac{1}{C_v-1}\quad\sum_{\mathclap{\h'\in \g_v\backslash \{\h \}}} \operatorname{M}_{\phi_{k}}\left(\h_{k}, \h_{k}' \right) \\
\h_{k+1} & =\operatorname{U}_{k}\parentheses*{\h_{k}, \boldsymbol{m}_{k+1}^\h }.
\end{aligned}
\end{equation}
We define $\operatorname{U}_{k}\parentheses*{\h, \boldsymbol{m}^\h} = \h + \boldsymbol{m}^\h$ and use a neural network for each $\operatorname{M}_{\phi_{k}}$. $\operatorname{M}_{\phi_{k}}$ acts on the concatenation of $\h_{k}$ and $\h_{k}'$.
Finally, the readout phase computes the final representation across the graph according to:
\begin{equation}
    \z^{\g_v} = \operatorname{R}_{\phi_{R}}\parentheses*{\frac{1}{C_v}\sum_{\h \in \g_v}\h_{K}},
\end{equation}
where $\operatorname{R}_{\phi_{R}}$ is a neural network. 

Since the same $\operatorname{M}_{\phi_{k}}$ is applied to all nodes at round $k$ and $\operatorname{R}_{\phi_{R}}$ simply operates on the mean across all final hidden states, the $\operatorname{MPNN}_{\phi}$ is able to compute the intra-graph representation on graphs of arbitrary sizes. This enables us to choose the size of the input graphs during pretraining randomly. Furthermore, for a downstream task with $C^{d}$ channels, we can simply produce the representations $[\h^1,\dots,\h^{C^d}]$, form a fully connected graph $\g$, and use the pretrained $\operatorname{MPNN}_{\phi}$ to compute one final representation $\z^\g$.

\subsection{Contrastive losses}
For each of the two settings, we pretrain a neural network with three different contrastive losses.
\\
\textbf{NT-Xent loss~\cite{simclr}}: Given a batch of $N$ samples consisting of $V$ views, the loss is computed pairwise for each pair of flattened views, $\mathbf{z}^v \in \mathbb{R}^{N\times L\cdot T_{\text{out}}}$ and  $\mathbf{z}^w\in \mathbb{R}^{N\times L \cdot T_{\text{out}}}$, also referred to as the positive pairs. The remaining $2N-1$ samples across both views are used as negative examples. Let $\cossit{i}{j}{w}{v} = \frac{\z_i^{w}\cdot\z_j^{v}}{\tau\|\z_i^{w}\|\cdot\|\z_j^{v}\|}$ denote the $\tau$-scaled cosine similarity between $\z_i^{w}$ and $\z_j^{v}$. The loss for one positive pair then becomes:
\begin{equation}
\small
\ell^{(w,v)}_{i}=\ln \frac{\exp\parentheses*{\cossit{i}{i}{w}{v}}}
{\sum\limits_{j}^{N} \exp\parentheses*{\cossit{i}{j}{w}{v}} + \sum\limits_{j\neq i}^{N} \exp\parentheses*{\cossit{i}{j}{w}{w}}}
\end{equation}
We compute $\ell^{(w,v)}$ for all positive pairs in the batch and average over them:
\begin{equation}
\small
   \mathcal{L}_{\text{NT-Xent}}^{(w,v)} = -\frac{1}{N}\sum_{i}^N \ell_i^{(w,v)}
\end{equation}
This operation is repeated for all combinations of views:
\begin{equation}
\small
    \mathcal{L}_{\text{NT-Xent}} = \frac{1}{ V(V-1)}\sum_{v}^{V}\sum_{w\neq v}^ V 
    \mathcal{L}^{(w,v)}_{\text{NT-Xent}}
\end{equation}
\\
\textbf{TS2Vec loss~\cite{TS2Vec}}: The TS2Vec loss also takes each combination of positive pairs. However, instead of flattening the tokens produced by the encoder, the loss takes the temporal relations in the representations into account. This is done by constructing two different versions of the negative examples and using these to compute a temporal loss and an instance loss, respectively. In the temporal loss, the negative examples are the remaining time stamps within the same sequence, $i$. In the instance loss, the negative examples are the remaining sequences in the batch at the same time stamp, $t$:
\begin{equation}
\small
{
\begin{aligned}
\ell t_{(i,t)}^{(w, v)}&=\ln \frac{\exp \left(\mathbf{z}_{i, t}^{w} \cdot \mathbf{z}_{i, t}^{v}\right)}{\sum\limits_{t^{\prime}}^{T_{\text{out}}}\exp \left(\mathbf{z}_{i, t}^{w} \cdot \mathbf{z}_{i, t^{\prime}}^{v}\right)+\sum\limits_{t^{\prime}\neq t}^{T_{\text{out}}}\exp \left(\mathbf{z}^{w}_{i, t} \cdot \mathbf{z}^{w}_{i, t^{\prime}}\right)} \\
\ell i_{(i,t)}^{(w,v)}&=\ln \frac{\exp \left(\mathbf{z}_{i, t}^{w} \cdot \mathbf{z}_{i, t}^{v}\right)}{\sum\limits_{j}^N\left(\exp \mathbf{z}^{w}_{i, t} \cdot \mathbf{z}_{j, t}^{v}\right)+\sum\limits_{j\neq i}^N \exp \left(\mathbf{z}^{w}_{i, t} \cdot \mathbf{z}^{w}_{j, t}\right)} \\
\end{aligned}}
\end{equation}
The temporal loss and instance loss are added to form the dual loss, $\mathcal{L}^{(w,v)}_{\text{dual}}=-\frac{1}{2 N T} \sum\limits_{i}^{N} \sum\limits_t^{T_{\text{out}}}\left(\ell t_{(i,t)}^{(w, v)}+\ell i_{(i,t)}^{(w,v)}\right)$.
The loss is then computed hierarchically by iteratively applying a maxpool operation across the temporal dimension of the representations and recomputing the dual loss to form $\mathcal{L}^{(w,v)}_{\text{TS2Vec}}$~\cite{TS2Vec}. Finally, this loss is also computed for all combinations of views:
\begin{equation}
\small
    \mathcal{L}_{\text{TS2Vec}} = \frac{1}{ V\cdot(V-1)}\sum_{v}^{V}\sum_{w\neq v}^ V \mathcal{L}_{\text{TS2Vec}}^{(w,v)}.
\end{equation}
\textbf{COCOA loss~\cite{cocoa}}: The COCOA loss is meant to reduce the computational complexity associated with NT-Xent when contrasting more than two views and also acts on flattened versions of $\z$. 
The loss separately computes the cross-view correlation (i.e., correlation between the positive pairs) as:
\begin{equation}
\small
\mathcal{L}_{\text{C}}^i =\sum_{v}^V \sum_{w\neq v}^V\fexp{1/\tau-\cossit{i}{i}{w}{v}}
\end{equation}
and the intra-view discriminator. The intra-view discriminator computes the correlation between the negative examples. The negative examples are only taken from the corresponding view, $v$, of the remaining examples in the batch:
\begin{equation}
\small
\mathcal{L}_{\text{D}}^v=\frac{1}{N} \sum_{i}^N\sum_{j\neq i}^N \fexp{\cossit{i}{j}{v}{v}}.
\end{equation}
The cross-view correlation and intra-view discriminator are then combined into the final loss:
\begin{equation}
\small
\mathcal{L}_{\text{COCOA}}=\sum_{i}^N \mathcal{L}_{\text{C}}^i+\lambda \sum_{v}^V \mathcal{L}_{\text{D}}^v
\end{equation}

\section{Experimental setup}
The implementation is available at \url{https://github.com/theabrusch/Multiview_TS_SSL}.
\subsection{Data}
For pretraining, we use the Physionet Challenge 2018 (PC18) dataset \cite{yousnooze, PhysioNet}, which is a dataset annotated for sleep staging. We use the EEG data from the 994 subjects of the training set to pretrain the models. The dataset contains the following six EEG channels; F3-M2, F4-M1, C3-M2, C4-M1, O1-M2, and O2-M1. All of the data is resampled from 200 Hz to 100 Hz. We split the subjects 0.8/0.2 for training and validating and then segment the entire dataset into 30s windows with no overlap. This results in 710,942 windows for pretraining and 178,569 windows for tracking the validation performance. 

For fine-tuning, we use the SleepEDFx dataset~\cite{PhysioNet, sleepedf}. The dataset contains 153 nights of sleep recordings from 78 subjects, and the data is annotated for sleep staging. Sleep staging gives rise to the following five classes; wake, N1, N2, N3 and R (the last four are different phases of sleep). The aim is to predict the sleep stage for windows of length 30s. 
We use the EEG data, which contains two channels; Fpz-Cz and Pz-Oz, sampled at 100Hz. 
We split the subjects 0.6/0.2/0.2 for training, validating, and testing. The splits are kept fixed throughout all experiments. Again, we segment the data into 30s windows with no overlap, yielding 122,016 and 37,379 windows available for training and validation and 36,955 windows for testing. In practice, we downsample the number of training and validation windows to simulate a setting with only a few labels available for fine-tuning. This process is described in \autoref{sec:fine-tuning}.

All windows in the pretraining and fine-tuning datasets are standardized, so each channel has zero mean and a standard deviation of one. 

\subsection{Model architecture}
We follow~\cite{BENDR} and use 6 convolutional blocks consisting of a 1D convolution, a dropout layer, a group normalization layer, and a GELU activation function. The kernel width is 3 in the first layer and 2 in the remaining 5 layers, and the stride is set to the same value as the width. We use 256 kernels for all intermediate layers and set the output dimension of the final layer to 64. Finally, we add a readout layer with kernel width and stride set to 1. This gives an output dimension $\h^{c}\in\mathbb{R}^{N\times L=64\times T_{\text{out}}=33}$. 

For the MPNN, we use a single linear layer followed by a dropout layer and a ReLU activation layer for all $\operatorname{M}_{\phi_{k}}$. The linear layer only acts on the second dimension of $\h^{c}$, i.e. the same weights are applied at all time steps $t\in T_{\text{out}}$. Thus, since it takes in two hidden states at a time, the dimension of the weights are $2\cdot64\times64$. For $\operatorname{R}_{\phi_{R}}$, we use two linear layers separated by a dropout layer and a ReLU activation function. 
\subsection{Pretraining setup}
During pretraining, all of our models are trained for 10 epochs. We use the AdamW optimizer with a learning rate of $10^{-3}$ and a weight decay of $10^{-2}$. We apply a dropout rate of 10\% between all layers in the network. All of the pretrained models are trained using a batch size of 64.

We benchmark our results against BENDR~\cite{BENDR} and Seq\-CLR~\cite{seqclr}. For BENDR, we use the original hyperparameters with no additional fine-tuning. Since our input to the model is smaller than what was used in the original paper, it is likely that more optimal masking parameters exist. Following their code, the pretraining is stopped if the network learns to precisely reconstruct tokens. This happens in our version of BENDR after 5900 iterations (around halfway into the first epoch).

For SeqCLR, we pretrain a version of their recurrent neural network (SeqCLR\_R), since this is reported to show the best results on the sleep staging dataset. We pretrain on windows of size 30s since this yielded better results than the 20s reported in the paper. We adjust their augmentations to our sampling frequency and input size. 
All implementation details are in the Git repository. 

\subsection{Fine-tuning}
\label{sec:fine-tuning}
When fine-tuning without the MPNN, we use a linear layer across all $\z^{c}$ to obtain one representation $\z\in\mathbb{R}^{N\times64\times 33}$ for classification. With the MPNN setup, we simply use the pretrained $\operatorname{MPNN}_{\phi}$ to obtain one representation across all channels. 
Subsequently, we average pool along the time dimension to obtain $T=4$ and flatten the representation, i.e., $\z^{\text{final}}\in\mathbb{R}^{N\times4\cdot64=256}$. We then use a single linear layer followed by a softmax operation that classifies each window. 

All of our models are fine-tuned with a learning rate of $5\cdot 10^{-4}$. We use the AdamW optimizer with a weight decay of $10^{-2}$. The batch size is set to 32. As we are generally interested in settings with few labels available for fine-tuning, we test the model by sampling a balanced set from the full dataset available for fine-tuning. We sample 10, 25, 50, 100, 200, 500, and 1000 data points per class respectively, and compare the performance of each of the pretrained models with the same models trained from scratch. We sample the same number of data points from the validation set. All models are fine-tuned for a maximum of 40 epochs, using early stopping on the validation loss with a patience of 7 epochs. We do this for both of the following settings: one where we optimize the entire network during fine-tuning and one where we only optimize the final linear layer(s) during fine-tuning. 

Since the BENDR encoder acts on a fixed input dimension, it is less trivial to fine-tune on a dataset with a different set of input channels. We insert the channels of the fine-tuning dataset at the position of the closest channel in the pretraining dataset. Therefore, we insert the channel Fpz-Cz at the position of both the F3-M3 and the F4-M1 channels, and insert the channel Pz-Oz in the same position as O1-M1 and O2-M1. We insert zeros at the positions of the remaining channels. 

\section{Results and discussion}
We run the fine-tuning experiments for five different seeds (i.e., both the data sampling and model initialization are reseeded five times) and report the averaged scores. 

\begin{figure}[t]
    \centering
    \includegraphics[width = \linewidth]{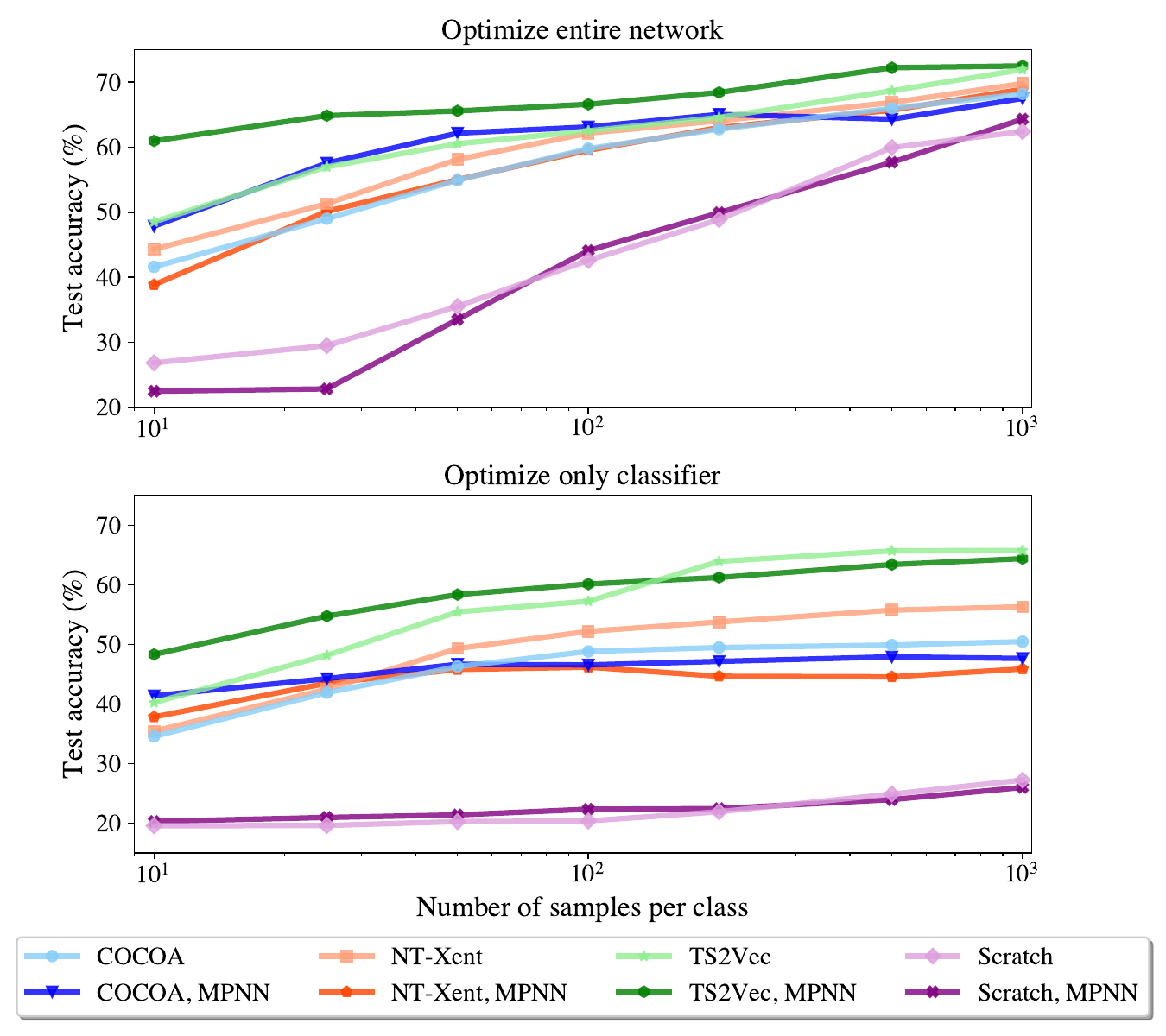}
    \caption{Balanced accuracy scores when optimizing the entire network (top) and freezing the encoder (bottom) during fine-tuning. Scores are averaged across 5 seeds.}
    \label{fig:optimize_encoder}
\end{figure}

\autoref{fig:optimize_encoder} (top) shows the results when optimizing the entire network during fine-tuning. \autoref{tab:optimize_encoder} shows a subset of the sample sizes and also includes the results from BENDR and the different versions of SeqCLR. All scores reported are balanced accuracy scores. Since it is a five-class problem, the chance level is 20\%. 
The table shows that the two networks trained from scratch yield similar results. However, the MPNN model has a lower score for fewer samples and higher scores for more samples compared to the non-MPNN model. This is likely due to the higher amount of trainable parameters in the MPNN model, which makes it more likely to overfit on small sample sizes. 
It is also clear that all of the pretraining schemes improve the score across all sample sizes. Both BENDR and SeqCLR\_R
improve the results over the models trained from scratch, but at a lower margin compared to the remaining models. We also pretrained a model with the same encoder architecture as our own models, but using the SeqCLR augmentations for contrastive learning. This model showed similar results as the SeqCLR\_R model, indicating that the multi-view pretraining strategy is beneficial when transferring between tasks with variable input channels. 

Comparing the pretrained MPNN models to the pretrained non-MPNN models, the picture is less clear. Where the MPNN model trained with the COCOA loss outperforms the non-MPNN model on smaller sample sizes, the MPNN+NT-Xent model performs worse than its non-MPNN counterpart on all sample sizes. Nonetheless, both \autoref{tab:optimize_encoder} and \autoref{fig:optimize_encoder} clearly demonstrate that the MPNN model trained with the TS2Vec loss outperforms all other pretraining schemes at all sample sizes. This is especially the case for the smallest sample size where the margin to the second highest score is 12.5\%.  
\begin{table}[t]
    \centering
    \footnotesize
    \begin{tabular}{@{}llrrrr@{}}
    \toprule
        Model & Pretraining & \multicolumn{4}{c}{ Samples per class} \\
        \cmidrule(lr){3-6}
                       &  & 10 & 50 & 100 & 1000 \\
        \midrule
        BENDR        & BENDR  & $.284$ & $.494$ & $.543$ & $.657$ \\ 
        SeqCLR\_R & NT-Xent & $.308$ & $.460$ & $.559$ & $.643$ \\
        SeqCLR\_R & Scratch & $.234$ & $.362$ &  $.398$ & $.564$ \\  \hline
Wo. MPNN & COCOA & $.416$ & $.549$ & $.598$ & $.682$ \\
Wo. MPNN & NT-Xent & $.443$ & $.581$ & $.622$ & $.698$ \\
Wo. MPNN & TS2Vec & $.485$ & $.605$ & $.625$ & $.719$ \\
Wo. MPNN & Scratch & $.268$ & $.355$ & $.426$ & $.624$ \\ \hline
W. MPNN & COCOA & $.479$ & $.622$ & $.631$ & $.675$ \\
W. MPNN & NT-Xent & $.388$ & $.550$ & $.595$ & $.689$ \\
W. MPNN & TS2Vec & $\mathbf{.610}$ & $\mathbf{.656}$ & $\mathbf{.666}$ & $\mathbf{.725}$ \\
W. MPNN & Scratch & $.225$ & $.335$ & $.441$ & $.643$ \\
        \bottomrule
    \end{tabular}
    \caption{Balanced accuracy scores after optimizing the entire network during fine-tuning averaged across 5 seeds.}
    \label{tab:optimize_encoder}
\end{table}
It therefore seems that the MPNN clearly improves the pretraining when combined with a loss that explicitly considers the temporal relation in the data. 


\begin{table}[h]
    \centering
    \footnotesize
    \begin{tabular}{@{}llrrrr@{}}
    \toprule
        Model & Pretraining & \multicolumn{4}{c}{ Samples per class} \\
        \cmidrule(lr){3-6}
                       &  & 10 & 50 & 100 & 1000 \\
        \midrule
        BENDR & BENDR  & $.201$ & $.215$ & $.227$ & $.263$\\ 
        SeqCLR\_R & NT-Xent & $.256$ & $.305$ & $.308$ & $.577$ \\
        SeqCLR\_R & Scratch & $.278$ & $.325$ &  $.338$ & $.363$ \\ \hline
        Wo. MPNN & COCOA & $.346$ & $.463$ & $.488$ & $.505$ \\
        Wo. MPNN & NT-Xent & $.355$ & $.493$ & $.522$ & $.563$ \\
        Wo. MPNN & TS2Vec & $.403$ & $.555$ & $.573$ & $\mathbf{.658}$ \\
        Wo. MPNN & Scratch & $.196$ & $.203$ & $.204$ & $.273$ \\ \hline
        W. MPNN & COCOA & $.414$ & $.467$ & $.466$ & $.477$ \\
        W. MPNN & NT-Xent & $.379$ & $.458$ & $.462$ & $.459$ \\
        W. MPNN & TS2Vec & $\mathbf{.483}$ & $\mathbf{.584}$ & $\mathbf{.601}$ & $.644$ \\
        W. MPNN & Scratch & $.203$ & $.214$ & $.224$ & $.260$ \\
        \bottomrule
    \end{tabular}
    \caption{Balanced accuracy scores after freezing the encoder and the MPNN during fine-tuning averaged across 5 seeds.}
    \label{tab:no_optimize_encoder}
\end{table}

\autoref{tab:no_optimize_encoder} and \autoref{fig:optimize_encoder} (bottom) show the results when freezing the encoder and MPNN during fine-tuning. Since we only optimize linear layers for these results, it is clear that the representations learned during pretraining for almost all models are transferable to a dataset with completely different channels. However, the BENDR results are comparable to the results of the randomly initialized models.
The table shows that while the MPNN+COCOA loss and MPNN+NT-Xent models achieve a slightly higher score than their non-MPNN counterparts on the smallest sample size, they perform worse on all other sample sizes. The MPNN+TS2Vec model again outperforms all other models for sample sizes smaller than 200 samples per class, whereas the non-MPNN+TS2Vec model achieves the highest performance for 200 samples or more per class.  

Thus, the results indicate that the pretrained MPNN helps in optimally combining the two channels for smaller sample sizes. When more data is available, the non-MPNN models are able to learn a better combination using the linear layer that is also optimized during fine-tuning. When fine-tuning on two channels, the complexity of inter-channel interactions is limited. We hypothesize that the pretrained MPNN is even more useful when fine-tuning on datasets with more than two channels and thus increased inter-channel complexity. 

Finally, we reiterate that the pretraining of BENDR was not optimized for our dataset. It is therefore likely that more optimal pretraining settings exist. However, the results still demonstrate the issue with existing pretraining schemes, where transferring between datasets with varying input channels is non-trivial. The methods presented here alleviate this issue. 

\section{Conclusions}
Self-supervised learning for multivariate time series suffers from the limitation that the input variables may vary from pretraining task to downstream task. Therefore, we proposed a channel-agnostic pretraining scheme applying the same encoder to all incoming channels and combining the channels using an MPNN. We compared our method to a network trained without the MPNN and the corresponding models with no pretraining and repeated our experiments for three different contrastive loss functions. We demonstrated the capability of the model by pretraining on a dataset with six EEG channels using a multi-view strategy for contrastive learning and fine-tuning on a dataset with two different EEG channels. We also compared to a model pretrained using augmentations for contrastive learning.

Our results showed that the MPNN model trained with a TS2Vec multi-view loss outperformed all other methods at all sample sizes when the entire network was optimized during fine-tuning. The same pattern was repeated when freezing the pretrained network during fine-tuning, although the MPNN+TS2Vec model was slightly outperformed by the non-MPNN+TS2Vec model at larger sample sizes. Our results demonstrated the potential of MPNNs combined with the multi-view strategy in creating a channel-agnostic pretraining scheme allowing for great flexibility when transferring between variable-channel datasets.

\bibliographystyle{IEEEbib}
\small
\bibliography{refs2}

\end{document}